\pdfoutput=1

\documentclass[11pt]{article}

\usepackage{acl}

\usepackage{times}
\usepackage{float}
\usepackage{latexsym}

\usepackage[T1]{fontenc}

\usepackage[utf8]{inputenc}

\usepackage{microtype}
\usepackage{booktabs}
\usepackage{hyperref}
\usepackage{multirow}

\usepackage{times}
\usepackage{latexsym}
\usepackage{graphicx}
\usepackage{amsmath}
\usepackage{amssymb}

\usepackage{booktabs}
\usepackage{colortbl}

\usepackage{soul}

\usepackage{comment}

\usepackage{balance}

\title{SemEval-2025 Task 7: Multilingual and Crosslingual Fact-Checked Claim Retrieval}

\author{Qiwei Peng$^{\heartsuit}$ \and Robert Moro$^{\spadesuit}$ \and Michal Gregor$^{\spadesuit}$ \and Ivan Srba$^{\spadesuit}$ \\ \textbf{Simon Ostermann}$^{\diamondsuit}$ \and \textbf{Marian Simko}$^{\spadesuit}$ \and \textbf{Juraj Podroužek}$^{\spadesuit}$ \\ \textbf{Matúš Mesarčík}$^{\spadesuit}$ \and \textbf{Jaroslav Kopčan}$^{\spadesuit}$ \and \textbf{Anders Søgaard}$^{\heartsuit}$  \\
$\heartsuit$University of Copenhagen \qquad $\spadesuit$Kempelen Institute of Intelligent Technologies \\ $\diamondsuit$German Research Institute for Artificial Intelligence (DFKI) \\
\{qipe, soegaard\}@di.ku.dk$^{\heartsuit}$ \qquad \{name.surname\}@kinit.sk$^{\spadesuit}$ \qquad simon.ostermann@dfki.de$^{\diamondsuit}$
}

\begin{document}
\maketitle

\begin{abstract}
The rapid spread of online disinformation presents a global challenge, and machine learning has been widely explored as a potential solution. However, multilingual settings and low-resource languages are often neglected in this field. To address this gap, we conducted a shared task on multilingual claim retrieval at SemEval 2025, aimed at identifying fact-checked claims that match newly encountered claims expressed in social media posts across different languages. The task includes two subtracks: (1) a monolingual track, where social posts and claims are in the same language, and (2) a crosslingual track, where social posts and claims might be in different languages. A total of 179 participants registered for the task contributing to 52 test submissions. 23 out of 31 teams have submitted their system papers. In this paper, we report the best-performing systems as well as the most common and the most effective approaches across both subtracks. This shared task, along with its dataset and participating systems, provides valuable insights into multilingual claim retrieval and automated fact-checking, supporting future research in this field.   

\end{abstract}

\section{Introduction}
The sheer amount of disinformation on the Internet is proving to be a societal problem~\cite{vosoughi2018spread}. Fact-checking organizations have made progress in manually and professionally fact-checking various contents \citep{vlachos2014fact, micallef2022true}. However, manual fact-checking at scale is too costly. To reduce some of the fact-checkers' manual efforts and make their work more effective, recent studies have examined their needs and identified tasks that could be automated \citep{ijcai2021p0619, zeng2021automated, hrckova2024autonomation}, such as evidence retrieval \citep{schuster2020limitations, liao2023muser} and verdict prediction \citep{nakashole2014language, thorne-etal-2018-fever}.

\begin{table}[t]
\small
    \centering
    \resizebox{\columnwidth}{!}{%
    \begin{tabular}{p{1cm}|p{3.5cm}|p{3.5cm}}
    \toprule
        & \textbf{Original text (German)} & \textbf{English translation }\\ \midrule 
        \textbf{Social media post} & \textit{Wer an Wahlen glaubt, ist auch der Meinung, dass das Ordnungsamt die Küche putzt.} Wussten Sie eigentlich das Das Besatzerstatut besagt: dass die Fremdverwaltung allein Durch die Wahlen vom Wahler akzeptiert wird ** Bei unter 50\% Wahlbeteiligung wäre dies hinfällig Denn es gabe dann ein rechtliches Problem Das Besatzerstatut wäre ungültig... & \textit{Anyone who believes in elections also believes that the regulatory office cleans the kitchen.} Did you actually know that? The occupier statute states: that the foreign administration is only accepted by the voters through the elections ** If the turnout is less than 50\%, this would be obsolete Because then there was a legal problem The occupier statute would be invalid...\\ \midrule
        \textbf{Claim} & Deutschland ist ein besetztes Land. & Germany is an occupied country.  \\
    \bottomrule
    \end{tabular}
    }
    \caption{An example from our dataset. The \textbf{social media post} was written by a user. The main text of the post is in \textit{italics}, the OCR transcript from an attached images is in regular font. The \textbf{claim} comes from the fact-check assigned by a fact-checker.}
    \label{tab:example}
\end{table}

In this work, we put focus on the task of \textit{claim retrieval}, also called \textit{claim matching}~\cite{kazemi-etal-2021-claim}, \textit{searching for fact-checked information}~\cite{vo-lee-2020-facts}, or \textit{fact-checked claim detection}~\cite{shaar-etal-2020-known}. Claim retrieval is an NLP task that addresses one significant problem that fact-checkers face: How to find out whether a similar claim has already been fact-checked before, even in a different language~\cite{hrckova2024autonomation}. Solving this problem would reduce duplicate work and improve the efficiency of fact-checking efforts. Previously, this task was mostly done in English. Other languages that have been considered include
Arabic, Bengali, Hindi, and Tamil \citep{kazemi-etal-2021-claim, nakov2022overview}. However, many other languages or even entire major language families have not been considered at all. 


To close this gap, we organized the SemEval 2025 Task 7 ``Multilingual and Crosslingual Fact-Checked Claim Retrieval''. We formulate claim retrieval task as an information retrieval task: Based on an input text (we use social media posts), the goal is to find an appropriate claim from a database of claims that have been already fact-checked by professional fact-checkers. Consider the example in Table~\ref{tab:example}. A user has written a post making a claim worth fact-checking. The idea of \textit{claim retrieval} is for a model to find a semantically similar claim from a list of previously fact-checked claims. We base the task on our already published multilingual dataset \textit{MultiClaim} \citep{pikuliak-etal-2023-multilingual} that consists of two parts: (1)~A list of 206k~fact-checked claims, and (2)~28k~social media posts (SMPs) with references to relevant fact-checked claims from the list. With these data, we can simulate a situation where a fact-checker is asked to fact-check an SMP, and they want to search through the list of already available fact-checks to see whether the same claim was fact-checked before. Our task features subtracks on both a monolingual and a crosslingual setup. To construct the training and development sets, we use a subset of the MultiClaim data set covering 8 different languages in the monolingual track and 14 different languages (52 combinations) in the crosslingual track. We further collect new resources to construct a test set which covers two additional previously unannounced languages and more than 4,000 new SMP-Claim pairs.

Our shared task attracted 179 participants. There were in total 52 test submissions by 31 teams, and 23 teams submitted system description papers. In this paper, we describe in detail the setup and implementation of our shared task along with datasets and submitted machine learning systems. These systems tackling the multilingual and crosslingual claim retrieval task provide valuable insights on applying state-of-the-art techniques to the real-world task and highlight future directions on better solving the claim retrieval problem.

\section{Related Work}
The increasing prevalence of disinformation on the Internet has prompted extensive research on fact-checking. A number of tasks have been proposed accordingly to examine different aspects of the process \citep{kotonya2020explainable, guo2022survey}, including claim detection \citep{Arslan_Hassan_Li_Tremayne_2020, 10.1145/3412869, eyuboglu2023fight}, verdict prediction \citep{nakashole2014language, thorne-etal-2018-fever, kumar2024verdict}, evidence retrieval \citep{schuster2020limitations, liao2023muser}, and justification production \citep{lu2020gcan, 10.1162/tacl_a_00601}. 

Prior shared tasks and competitions have touched on claim verification, such as the well-known FEVER shared task \citep{thorne-etal-2018-fact, fever-2020-fact, fever-2023-fact}, and AVeriTeC shared task \citep{schlichtkrull2023averitec, schlichtkrull-etal-2024-automated}. Evidence retrieval has also attracted significant attention. \citet{jullien-etal-2023-semeval} organized the shared task on evidence retrieval on clinical trial data. SemEval 2020 Task 9 \citep{wang-etal-2021-semeval} focuses on evidence finding for tabular data in scientific documents. Similarly, shared tasks are organized to tackle claim retrieval in social media posts, such as the CheckThat! Lab 2023 shared task \citep{10.1007/978-3-031-56069-9_62}, and in a multi-modal and multigenre content, such as CheckThat! Lab 2024 \citep{alam2023overview} and Factify 2 \citep{suryavardan2023findings}. 

\begin{table*}[]
\centering
\resizebox{0.65\linewidth}{!}{%
\begin{tabular}{|c|ccc|cc|ccc|}
\hline
\multirow{2}{*}{Language} & \multicolumn{3}{c|}{\# of posts} & \multicolumn{2}{c|}{\# of claims} & \multicolumn{3}{c|}{\# of pairs} \\
                      & train & dev & test & train \& dev & test  & train & dev & test  \\ \hline
Arabic                & 676   & 78    & 500    & 14,201 & 21,153  & 680    & 78  & 514   \\ \hline
English               & 4,351 & 478   & 500    & 85,734 & 145,287 & 5,446  & 627 & 574   \\ \hline
French                & 1,596 & 188   & 500    & 4,355  & 6,316   & 1,667  & 200 & 740   \\ \hline
German                & 667   & 83    & 500    & 4,996  & 7,485   & 830    & 101 & 549   \\ \hline
Malay                 & 1,062 & 105   & 93     & 8,424  & 686     & 1,169  & 116 & 96    \\ \hline
Portuguese            & 2,571 & 302   & 500    & 21,569 & 32,598  & 3,386  & 403 & 613   \\ \hline
Spanish               & 5,628 & 615   & 500    & 14,082 & 25,440  & 6,313  & 692 & 576   \\ \hline
Thai                  & 465   & 42    & 183    & 382    & 583     & 465    & 42  & 183   \\ \hline
Polish                & -     & -     & 500    & -      & 8,796   & -      & -   & 564   \\ \hline
Turkish               & -     & -     & 500    & -      & 12,536  & -      & -   & 550   \\ \hline\hline
Monolingual (total)  & 17,016 & 1,891 & 4,276 & 153,743 & 260,880 & 19,956 & 2,259  & 4,959 \\
\hline\hline
Crosslingual         & 4,972 & 552 & 4,000  & 153,743    & 272,447 & 5,787 & 651 & 5,322 \\ \hline
\end{tabular}%
}
\caption{The dataset statistics reporting the number of posts, claims, and pairs for the monolingual (shown also per individual languages) as well as for the crosslingual track.}
\label{tab:dataset_statistics}
\end{table*}

\begin{table}
\centering
\resizebox{1\columnwidth}{!}{%
\begin{tabular}{lrrrrrrrr}
\toprule
Claim language & ara & deu & eng & fra & msa & por & spa & tha \\
SMP language &  &  &  &  &  &  &  &  \\
\midrule
Arabic (ara) & {\cellcolor[HTML]{E4EFF9}} \color[HTML]{000000} 94 & {\cellcolor[HTML]{F7FBFF}} \color[HTML]{000000} {\cellcolor{white}} 0 & {\cellcolor[HTML]{F4F9FE}} \color[HTML]{000000} 17 & {\cellcolor[HTML]{F4F9FE}} \color[HTML]{000000} 19 & {\cellcolor[HTML]{F7FBFF}} \color[HTML]{000000} 3 & {\cellcolor[HTML]{F7FBFF}} \color[HTML]{000000} {\cellcolor{white}} 0 & {\cellcolor[HTML]{F7FBFF}} \color[HTML]{000000} 2 & {\cellcolor[HTML]{F7FBFF}} \color[HTML]{000000} {\cellcolor{white}} 0 \\
German (deu) & {\cellcolor[HTML]{F7FBFF}} \color[HTML]{000000} 1 & {\cellcolor[HTML]{E7F0FA}} \color[HTML]{000000} 84 & {\cellcolor[HTML]{F6FAFF}} \color[HTML]{000000} {\cellcolor{white}} 4 & {\cellcolor[HTML]{F7FBFF}} \color[HTML]{000000} 3 & {\cellcolor[HTML]{F7FBFF}} \color[HTML]{000000} {\cellcolor{white}} 0 & {\cellcolor[HTML]{F6FAFF}} \color[HTML]{000000} 4 & {\cellcolor[HTML]{F4F9FE}} \color[HTML]{000000} 19 & {\cellcolor[HTML]{F7FBFF}} \color[HTML]{000000} 1 \\
English (eng) & {\cellcolor[HTML]{F5F9FE}} \color[HTML]{000000} 15 & {\cellcolor[HTML]{EEF5FC}} \color[HTML]{000000} 50 & {\cellcolor[HTML]{2F7FBC}} \color[HTML]{F1F1F1} 699 & {\cellcolor[HTML]{E5EFF9}} \color[HTML]{000000} 90 & {\cellcolor[HTML]{E7F1FA}} \color[HTML]{000000} 82 & {\cellcolor[HTML]{DCEAF6}} \color[HTML]{000000} 135 & {\cellcolor[HTML]{CBDEF1}} \color[HTML]{000000} 225 & {\cellcolor[HTML]{F4F9FE}} \color[HTML]{000000} 17 \\
French (fra) & {\cellcolor[HTML]{F6FAFF}} \color[HTML]{000000} 5 & {\cellcolor[HTML]{F7FBFF}} \color[HTML]{000000} {\cellcolor{white}} 0 & {\cellcolor[HTML]{F5F9FE}} \color[HTML]{000000} 13 & {\cellcolor[HTML]{CDDFF1}} \color[HTML]{000000} 218 & {\cellcolor[HTML]{F7FBFF}} \color[HTML]{000000} 2 & {\cellcolor[HTML]{F5F9FE}} \color[HTML]{000000} 12 & {\cellcolor[HTML]{F6FAFF}} \color[HTML]{000000} 6 & {\cellcolor[HTML]{F7FBFF}} \color[HTML]{000000} {\cellcolor{white}} 0 \\
Hindi (hin) & {\cellcolor[HTML]{F7FBFF}} \color[HTML]{000000} {\cellcolor{white}} 0 & {\cellcolor[HTML]{F7FBFF}} \color[HTML]{000000} {\cellcolor{white}} 0 & {\cellcolor[HTML]{083979}} \color[HTML]{F1F1F1} 964 & {\cellcolor[HTML]{F7FBFF}} \color[HTML]{000000} {\cellcolor{white}} 0 & {\cellcolor[HTML]{F7FBFF}} \color[HTML]{000000} {\cellcolor{white}} 0 & {\cellcolor[HTML]{F7FBFF}} \color[HTML]{000000} {\cellcolor{white}} 0 & {\cellcolor[HTML]{F7FBFF}} \color[HTML]{000000} {\cellcolor{white}} 0 & {\cellcolor[HTML]{F7FBFF}} \color[HTML]{000000} {\cellcolor{white}} 0 \\
Korean (kor) & {\cellcolor[HTML]{F7FBFF}} \color[HTML]{000000} {\cellcolor{white}} 0 & {\cellcolor[HTML]{F7FBFF}} \color[HTML]{000000} {\cellcolor{white}} 0 & {\cellcolor[HTML]{C4DAEE}} \color[HTML]{000000} 257 & {\cellcolor[HTML]{F7FBFF}} \color[HTML]{000000} {\cellcolor{white}} 0 & {\cellcolor[HTML]{F7FBFF}} \color[HTML]{000000} {\cellcolor{white}} 0 & {\cellcolor[HTML]{F7FBFF}} \color[HTML]{000000} {\cellcolor{white}} 0 & {\cellcolor[HTML]{F7FBFF}} \color[HTML]{000000} {\cellcolor{white}} 0 & {\cellcolor[HTML]{F7FBFF}} \color[HTML]{000000} {\cellcolor{white}} 0 \\
Malay (msa) & {\cellcolor[HTML]{F7FBFF}} \color[HTML]{000000} {\cellcolor{white}} 0 & {\cellcolor[HTML]{F7FBFF}} \color[HTML]{000000} 1 & {\cellcolor[HTML]{E4EFF9}} \color[HTML]{000000} 96 & {\cellcolor[HTML]{F7FBFF}} \color[HTML]{000000} {\cellcolor{white}} 0 & {\cellcolor[HTML]{DCEAF6}} \color[HTML]{000000} 135 & {\cellcolor[HTML]{F7FBFF}} \color[HTML]{000000} {\cellcolor{white}} 0 & {\cellcolor[HTML]{F7FBFF}} \color[HTML]{000000} {\cellcolor{white}} 0 & {\cellcolor[HTML]{F7FBFF}} \color[HTML]{000000} 3 \\
Portuguese (por) & {\cellcolor[HTML]{F7FBFF}} \color[HTML]{000000} 1 & {\cellcolor[HTML]{F7FBFF}} \color[HTML]{000000} {\cellcolor{white}} 0 & {\cellcolor[HTML]{F3F8FE}} \color[HTML]{000000} 22 & {\cellcolor[HTML]{F7FBFF}} \color[HTML]{000000} 3 & {\cellcolor[HTML]{F7FBFF}} \color[HTML]{000000} 2 & {\cellcolor[HTML]{97C6DF}} \color[HTML]{000000} 392 & {\cellcolor[HTML]{F0F6FD}} \color[HTML]{000000} 36 & {\cellcolor[HTML]{F7FBFF}} \color[HTML]{000000} 1 \\
Sinhala (sin) & {\cellcolor[HTML]{F7FBFF}} \color[HTML]{000000} {\cellcolor{white}} 0 & {\cellcolor[HTML]{F7FBFF}} \color[HTML]{000000} {\cellcolor{white}} 0 & {\cellcolor[HTML]{69ADD5}} \color[HTML]{F1F1F1} 504 & {\cellcolor[HTML]{F7FBFF}} \color[HTML]{000000} {\cellcolor{white}} 0 & {\cellcolor[HTML]{F7FBFF}} \color[HTML]{000000} {\cellcolor{white}} 0 & {\cellcolor[HTML]{F7FBFF}} \color[HTML]{000000} {\cellcolor{white}} 0 & {\cellcolor[HTML]{F7FBFF}} \color[HTML]{000000} {\cellcolor{white}} 0 & {\cellcolor[HTML]{F7FBFF}} \color[HTML]{000000} {\cellcolor{white}} 0 \\
Spanish (spa) & {\cellcolor[HTML]{F7FBFF}} \color[HTML]{000000} 3 & {\cellcolor[HTML]{F7FBFF}} \color[HTML]{000000} 1 & {\cellcolor[HTML]{EFF6FC}} \color[HTML]{000000} 40 & {\cellcolor[HTML]{F6FAFF}} \color[HTML]{000000} 7 & {\cellcolor[HTML]{F7FBFF}} \color[HTML]{000000} 2 & {\cellcolor[HTML]{F0F6FD}} \color[HTML]{000000} 36 & {\cellcolor[HTML]{1663AA}} \color[HTML]{F1F1F1} 804 & {\cellcolor[HTML]{F7FBFF}} \color[HTML]{000000} {\cellcolor{white}} 0 \\
Tagalog (tgl) & {\cellcolor[HTML]{F7FBFF}} \color[HTML]{000000} {\cellcolor{white}} 0 & {\cellcolor[HTML]{F7FBFF}} \color[HTML]{000000} {\cellcolor{white}} 0 & {\cellcolor[HTML]{C3DAEE}} \color[HTML]{000000} 258 & {\cellcolor[HTML]{F7FBFF}} \color[HTML]{000000} {\cellcolor{white}} 0 & {\cellcolor[HTML]{F7FBFF}} \color[HTML]{000000} {\cellcolor{white}} 0 & {\cellcolor[HTML]{F7FBFF}} \color[HTML]{000000} {\cellcolor{white}} 0 & {\cellcolor[HTML]{F7FBFF}} \color[HTML]{000000} {\cellcolor{white}} 0 & {\cellcolor[HTML]{F7FBFF}} \color[HTML]{000000} {\cellcolor{white}} 0 \\
Thai (tha) & {\cellcolor[HTML]{F7FBFF}} \color[HTML]{000000} {\cellcolor{white}} 0 & {\cellcolor[HTML]{F7FBFF}} \color[HTML]{000000} {\cellcolor{white}} 0 & {\cellcolor[HTML]{E6F0F9}} \color[HTML]{000000} 88 & {\cellcolor[HTML]{F7FBFF}} \color[HTML]{000000} {\cellcolor{white}} 0 & {\cellcolor[HTML]{F7FBFF}} \color[HTML]{000000} {\cellcolor{white}} 0 & {\cellcolor[HTML]{F7FBFF}} \color[HTML]{000000} {\cellcolor{white}} 0 & {\cellcolor[HTML]{F7FBFF}} \color[HTML]{000000} {\cellcolor{white}} 0 & {\cellcolor[HTML]{EEF5FC}} \color[HTML]{000000} 50 \\
Urdu (urd) & {\cellcolor[HTML]{F7FBFF}} \color[HTML]{000000} {\cellcolor{white}} 0 & {\cellcolor[HTML]{F7FBFF}} \color[HTML]{000000} {\cellcolor{white}} 0 & {\cellcolor[HTML]{9FCAE1}} \color[HTML]{000000} 373 & {\cellcolor[HTML]{F7FBFF}} \color[HTML]{000000} {\cellcolor{white}} 0 & {\cellcolor[HTML]{F7FBFF}} \color[HTML]{000000} {\cellcolor{white}} 0 & {\cellcolor[HTML]{F7FBFF}} \color[HTML]{000000} {\cellcolor{white}} 0 & {\cellcolor[HTML]{F7FBFF}} \color[HTML]{000000} {\cellcolor{white}} 0 & {\cellcolor[HTML]{F7FBFF}} \color[HTML]{000000} {\cellcolor{white}} 0 \\
Chinese (zho) & {\cellcolor[HTML]{F7FBFF}} \color[HTML]{000000} {\cellcolor{white}} 0 & {\cellcolor[HTML]{F7FBFF}} \color[HTML]{000000} {\cellcolor{white}} 0 & {\cellcolor[HTML]{5FA6D1}} \color[HTML]{F1F1F1} 539 & {\cellcolor[HTML]{F7FBFF}} \color[HTML]{000000} {\cellcolor{white}} 0 & {\cellcolor[HTML]{F7FBFF}} \color[HTML]{000000} {\cellcolor{white}} 0 & {\cellcolor[HTML]{F7FBFF}} \color[HTML]{000000} {\cellcolor{white}} 0 & {\cellcolor[HTML]{F7FBFF}} \color[HTML]{000000} {\cellcolor{white}} 0 & {\cellcolor[HTML]{F7FBFF}} \color[HTML]{000000} {\cellcolor{white}} 0 \\
\bottomrule
\end{tabular}
}
\caption{Combinations of languages in SMP-Claim pairs for train and dev sets of the crosslingual track.}
\label{tab:crosslingual_train_and_dev}
\end{table}

\section{Task Description}
The shared task is formulated as an information retrieval task. Given social media posts, the goal is to retrieve the most relevant claims from a list of claims that are previously fact-checked. We set up two tracks for the shared task. In the \textbf{monolingual track}, matched posts and claims are always in the same language and the search pools for candidate claims are language specific to the post. In the \textbf{crosslingual track}, the search pool of candidate claims is not restricted and the matched claims can be in languages that are different from the post.

\subsection{Dataset}
The dataset used in the shared task is available for research purposes at Zenodo\footnote{\url{https://doi.org/10.5281/zenodo.14989176}}. The training and development sets are based on the MultiClaim \citep{pikuliak-etal-2023-multilingual} dataset. The test set contains additional data, which are not part of the original MultiClaim dataset, but which were collected using the same methodology.

\paragraph{Training and Development Sets.} 
The original MultiClaim dataset consists of two parts:

1. \textit{Fact-checked claims}. The dataset contains 205,751 fact-checked claims extracted from fact-checks created by 142 fact-checking organizations. Claims are usually one sentence long summaries of the main idea that is being fact-checked.

2. \textit{Social media posts}. The dataset contains 28,092 SMPs spanning 27 different languages, collected from Facebook, Instagram, and Twitter by following the links to these platforms given in the fact-checking articles. This way, 31,305 SMP-Claim pairs (each SMP has at least one claim assigned) were established. All posts in MultiClaim were published before 2023.

To make sure that only relevant SMPs are considered, two filtering techniques were used: (1) We use links from the \texttt{ClaimReview} json schema that the fact-checkers use. (2) We use the fact-checking warnings provided by the Facebook and Instagram platforms. When a visitor visits SMPs that were flagged as disinformation, the warnings contain a link to relevant fact-checks. In both cases, we can be sure that the link between a claim and an SMP is correct, because a fact-checker left an explicit signal confirming it. Manual inspections of the connections (done by three authors on a random sample of 100 pairs; see ~\citealp{pikuliak-etal-2023-multilingual} for more details) confirmed the absence of false positives. However, roughly 15\% of the connections rely on visual information present in either attached images or videos, making it impossible to match the SMP with an appropriate claim based on the text data only.

An additional manual inspection (of claims retrieved for a sample of 87 posts done by three authors; see ~\citealp{pikuliak-etal-2023-multilingual} for more details) also revealed that there are many potential connections between SMPs and claims that are missing in the dataset due to the employed collection procedure, especially crosslingual connections. To at least partially mitigate this, we added to the dataset additional SMP-Claim pairs created by following \textit{transitive connections}: If two SMPs $p_i$ and $p_j$ share a link to the same fact-checked claim and one of those ($p_i$) is also linked to a different claim $c_k$, we add the link $(p_j, c_k)$, if missing. This way, we were able to identify 3,351 additional SMP-Claim pairs.

To construct high-quality training and development sets covering different languages, we filter out languages that have less than 500 SMP-Claim pairs. After filtering, the dataset contains 8 languages (Arabic, English, French, German, Malay, Portuguese, Spanish, and Thai) for the monolingual track. For the crosslingual track, we include all SMP-Claim pairs, where the language of the claim is different than the language of the post and at the same time, both of the languages are already included in the monolingual pairs; additionally, all pairs with a language combination  appearing at least 200 times are included even if these languages do not appear in the monolingual pairs. There are still only 8 claim languages, but there are 6 additional post languages as compared to the monolingual track (Hindi, Korean, Sinhala, Tagalog, Urdu, and Chinese). There is a total of 52 different combinations that are covered in the crosslingual track. To prevent an exploitation of the data design in the crosslingual task (e.g., by ignoring the language of the post, if the matched claim is always in a different language), we assign a randomly selected 10\% of monolingual pairs into the crosslingual subset of the data. The development set was created by holding out a randomly selected 10\% of pairs for the monolingual and the crosslingual tracks. The final numbers of selected posts, claims, and pairs are reported in Table~\ref{tab:dataset_statistics}. Table~\ref{tab:crosslingual_train_and_dev} shows the combinations of languages in SMP-Claim pairs for the combined training and development sets.

\begin{table}
\centering
\resizebox{1\columnwidth}{!}{
\begin{tabular}{lrrrrrrrrrrr}
\toprule
Claim language & ara & deu & eng & fra & hin & msa & pol & por & spa & tha & tur \\
SMP language &  &  &  &  &  &  &  &  &  &  &  \\
\midrule
Arabic (ara) & {\cellcolor[HTML]{CBDEF1}} \color[HTML]{000000} 226 & {\cellcolor[HTML]{F7FBFF}} \color[HTML]{000000} 2 & {\cellcolor[HTML]{F0F6FD}} \color[HTML]{000000} 39 & {\cellcolor[HTML]{F6FAFF}} \color[HTML]{000000} 4 & {\cellcolor[HTML]{F7FBFF}} \color[HTML]{000000} {\cellcolor{white}} 0 & {\cellcolor[HTML]{F7FBFF}} \color[HTML]{000000} {\cellcolor{white}} 0 & {\cellcolor[HTML]{F7FBFF}} \color[HTML]{000000} {\cellcolor{white}} 0 & {\cellcolor[HTML]{F7FBFF}} \color[HTML]{000000} 1 & {\cellcolor[HTML]{F5FAFE}} \color[HTML]{000000} 8 & {\cellcolor[HTML]{F7FBFF}} \color[HTML]{000000} {\cellcolor{white}} 0 & {\cellcolor[HTML]{F7FBFF}} \color[HTML]{000000} 3 \\
Bengali (ben) & {\cellcolor[HTML]{F7FBFF}} \color[HTML]{000000} {\cellcolor{white}} 0 & {\cellcolor[HTML]{F7FBFF}} \color[HTML]{000000} {\cellcolor{white}} 0 & {\cellcolor[HTML]{D9E8F5}} \color[HTML]{000000} 149 & {\cellcolor[HTML]{F7FBFF}} \color[HTML]{000000} {\cellcolor{white}} 0 & {\cellcolor[HTML]{F7FBFF}} \color[HTML]{000000} {\cellcolor{white}} 0 & {\cellcolor[HTML]{F7FBFF}} \color[HTML]{000000} {\cellcolor{white}} 0 & {\cellcolor[HTML]{F7FBFF}} \color[HTML]{000000} {\cellcolor{white}} 0 & {\cellcolor[HTML]{F7FBFF}} \color[HTML]{000000} {\cellcolor{white}} 0 & {\cellcolor[HTML]{F7FBFF}} \color[HTML]{000000} {\cellcolor{white}} 0 & {\cellcolor[HTML]{F7FBFF}} \color[HTML]{000000} {\cellcolor{white}} 0 & {\cellcolor[HTML]{F7FBFF}} \color[HTML]{000000} {\cellcolor{white}} 0 \\
German (deu) & {\cellcolor[HTML]{F7FBFF}} \color[HTML]{000000} {\cellcolor{white}} 0 & {\cellcolor[HTML]{E1EDF8}} \color[HTML]{000000} 110 & {\cellcolor[HTML]{F4F9FE}} \color[HTML]{000000} 16 & {\cellcolor[HTML]{F7FBFF}} \color[HTML]{000000} 2 & {\cellcolor[HTML]{F7FBFF}} \color[HTML]{000000} {\cellcolor{white}} 0 & {\cellcolor[HTML]{F7FBFF}} \color[HTML]{000000} {\cellcolor{white}} 0 & {\cellcolor[HTML]{F7FBFF}} \color[HTML]{000000} 2 & {\cellcolor[HTML]{F7FBFF}} \color[HTML]{000000} {\cellcolor{white}} 0 & {\cellcolor[HTML]{F6FAFF}} \color[HTML]{000000} 4 & {\cellcolor[HTML]{F7FBFF}} \color[HTML]{000000} {\cellcolor{white}} 0 & {\cellcolor[HTML]{F6FAFF}} \color[HTML]{000000} 5 \\
English (eng) & {\cellcolor[HTML]{F5F9FE}} \color[HTML]{000000} 14 & {\cellcolor[HTML]{EDF4FC}} \color[HTML]{000000} 51 & {\cellcolor[HTML]{08306B}} \color[HTML]{F1F1F1} 1,030 & {\cellcolor[HTML]{F2F7FD}} \color[HTML]{000000} 29 & {\cellcolor[HTML]{D3E3F3}} \color[HTML]{000000} 186 & {\cellcolor[HTML]{F6FAFF}} \color[HTML]{000000} 4 & {\cellcolor[HTML]{F5F9FE}} \color[HTML]{000000} 12 & {\cellcolor[HTML]{F2F8FD}} \color[HTML]{000000} 24 & {\cellcolor[HTML]{F0F6FD}} \color[HTML]{000000} 39 & {\cellcolor[HTML]{F7FBFF}} \color[HTML]{000000} 2 & {\cellcolor[HTML]{EEF5FC}} \color[HTML]{000000} 49 \\
French (fra) & {\cellcolor[HTML]{F7FBFF}} \color[HTML]{000000} 1 & {\cellcolor[HTML]{F6FAFF}} \color[HTML]{000000} 4 & {\cellcolor[HTML]{F2F7FD}} \color[HTML]{000000} 28 & {\cellcolor[HTML]{ECF4FB}} \color[HTML]{000000} 56 & {\cellcolor[HTML]{F7FBFF}} \color[HTML]{000000} {\cellcolor{white}} 0 & {\cellcolor[HTML]{F7FBFF}} \color[HTML]{000000} {\cellcolor{white}} 0 & {\cellcolor[HTML]{F7FBFF}} \color[HTML]{000000} {\cellcolor{white}} 0 & {\cellcolor[HTML]{F7FBFF}} \color[HTML]{000000} 1 & {\cellcolor[HTML]{F7FBFF}} \color[HTML]{000000} 3 & {\cellcolor[HTML]{F7FBFF}} \color[HTML]{000000} {\cellcolor{white}} 0 & {\cellcolor[HTML]{F7FBFF}} \color[HTML]{000000} 1 \\
Hindi (hin) & {\cellcolor[HTML]{F7FBFF}} \color[HTML]{000000} {\cellcolor{white}} 0 & {\cellcolor[HTML]{F7FBFF}} \color[HTML]{000000} {\cellcolor{white}} 0 & {\cellcolor[HTML]{08306B}} \color[HTML]{F1F1F1} 2,337 & {\cellcolor[HTML]{F7FBFF}} \color[HTML]{000000} {\cellcolor{white}} 0 & {\cellcolor[HTML]{F7FBFF}} \color[HTML]{000000} {\cellcolor{white}} 0 & {\cellcolor[HTML]{F7FBFF}} \color[HTML]{000000} {\cellcolor{white}} 0 & {\cellcolor[HTML]{F7FBFF}} \color[HTML]{000000} {\cellcolor{white}} 0 & {\cellcolor[HTML]{F7FBFF}} \color[HTML]{000000} {\cellcolor{white}} 0 & {\cellcolor[HTML]{F7FBFF}} \color[HTML]{000000} {\cellcolor{white}} 0 & {\cellcolor[HTML]{F7FBFF}} \color[HTML]{000000} {\cellcolor{white}} 0 & {\cellcolor[HTML]{F7FBFF}} \color[HTML]{000000} {\cellcolor{white}} 0 \\
Malay (msa) & {\cellcolor[HTML]{F7FBFF}} \color[HTML]{000000} {\cellcolor{white}} 0 & {\cellcolor[HTML]{F7FBFF}} \color[HTML]{000000} {\cellcolor{white}} 0 & {\cellcolor[HTML]{F5FAFE}} \color[HTML]{000000} 8 & {\cellcolor[HTML]{F7FBFF}} \color[HTML]{000000} {\cellcolor{white}} 0 & {\cellcolor[HTML]{F7FBFF}} \color[HTML]{000000} {\cellcolor{white}} 0 & {\cellcolor[HTML]{F6FAFF}} \color[HTML]{000000} 5 & {\cellcolor[HTML]{F7FBFF}} \color[HTML]{000000} {\cellcolor{white}} 0 & {\cellcolor[HTML]{F7FBFF}} \color[HTML]{000000} {\cellcolor{white}} 0 & {\cellcolor[HTML]{F7FBFF}} \color[HTML]{000000} {\cellcolor{white}} 0 & {\cellcolor[HTML]{F7FBFF}} \color[HTML]{000000} {\cellcolor{white}} 0 & {\cellcolor[HTML]{F7FBFF}} \color[HTML]{000000} {\cellcolor{white}} 0 \\
Polish (pol) & {\cellcolor[HTML]{F7FBFF}} \color[HTML]{000000} {\cellcolor{white}} 0 & {\cellcolor[HTML]{F7FBFF}} \color[HTML]{000000} {\cellcolor{white}} 0 & {\cellcolor[HTML]{F5F9FE}} \color[HTML]{000000} 14 & {\cellcolor[HTML]{F7FBFF}} \color[HTML]{000000} 2 & {\cellcolor[HTML]{F7FBFF}} \color[HTML]{000000} {\cellcolor{white}} 0 & {\cellcolor[HTML]{F7FBFF}} \color[HTML]{000000} {\cellcolor{white}} 0 & {\cellcolor[HTML]{ECF4FB}} \color[HTML]{000000} 55 & {\cellcolor[HTML]{F7FBFF}} \color[HTML]{000000} 1 & {\cellcolor[HTML]{F7FBFF}} \color[HTML]{000000} {\cellcolor{white}} 0 & {\cellcolor[HTML]{F7FBFF}} \color[HTML]{000000} {\cellcolor{white}} 0 & {\cellcolor[HTML]{F7FBFF}} \color[HTML]{000000} {\cellcolor{white}} 0 \\
Portuguese (por) & {\cellcolor[HTML]{F6FAFF}} \color[HTML]{000000} 6 & {\cellcolor[HTML]{F7FBFF}} \color[HTML]{000000} 3 & {\cellcolor[HTML]{F3F8FE}} \color[HTML]{000000} 21 & {\cellcolor[HTML]{F7FBFF}} \color[HTML]{000000} 2 & {\cellcolor[HTML]{F7FBFF}} \color[HTML]{000000} {\cellcolor{white}} 0 & {\cellcolor[HTML]{F7FBFF}} \color[HTML]{000000} {\cellcolor{white}} 0 & {\cellcolor[HTML]{F7FBFF}} \color[HTML]{000000} {\cellcolor{white}} 0 & {\cellcolor[HTML]{E8F1FA}} \color[HTML]{000000} 77 & {\cellcolor[HTML]{F5F9FE}} \color[HTML]{000000} 15 & {\cellcolor[HTML]{F7FBFF}} \color[HTML]{000000} {\cellcolor{white}} 0 & {\cellcolor[HTML]{F7FBFF}} \color[HTML]{000000} 2 \\
Spanish (spa) & {\cellcolor[HTML]{F7FBFF}} \color[HTML]{000000} 1 & {\cellcolor[HTML]{F7FBFF}} \color[HTML]{000000} {\cellcolor{white}} 0 & {\cellcolor[HTML]{EFF6FC}} \color[HTML]{000000} 40 & {\cellcolor[HTML]{F7FBFF}} \color[HTML]{000000} 1 & {\cellcolor[HTML]{F7FBFF}} \color[HTML]{000000} {\cellcolor{white}} 0 & {\cellcolor[HTML]{F7FBFF}} \color[HTML]{000000} {\cellcolor{white}} 0 & {\cellcolor[HTML]{F7FBFF}} \color[HTML]{000000} {\cellcolor{white}} 0 & {\cellcolor[HTML]{F5FAFE}} \color[HTML]{000000} 11 & {\cellcolor[HTML]{CDE0F1}} \color[HTML]{000000} 211 & {\cellcolor[HTML]{F7FBFF}} \color[HTML]{000000} {\cellcolor{white}} 0 & {\cellcolor[HTML]{F7FBFF}} \color[HTML]{000000} 3 \\
Thai (tha) & {\cellcolor[HTML]{F7FBFF}} \color[HTML]{000000} {\cellcolor{white}} 0 & {\cellcolor[HTML]{F7FBFF}} \color[HTML]{000000} {\cellcolor{white}} 0 & {\cellcolor[HTML]{F4F9FE}} \color[HTML]{000000} 18 & {\cellcolor[HTML]{F7FBFF}} \color[HTML]{000000} {\cellcolor{white}} 0 & {\cellcolor[HTML]{F7FBFF}} \color[HTML]{000000} {\cellcolor{white}} 0 & {\cellcolor[HTML]{F7FBFF}} \color[HTML]{000000} {\cellcolor{white}} 0 & {\cellcolor[HTML]{F7FBFF}} \color[HTML]{000000} {\cellcolor{white}} 0 & {\cellcolor[HTML]{F7FBFF}} \color[HTML]{000000} {\cellcolor{white}} 0 & {\cellcolor[HTML]{F7FBFF}} \color[HTML]{000000} {\cellcolor{white}} 0 & {\cellcolor[HTML]{F5F9FE}} \color[HTML]{000000} 13 & {\cellcolor[HTML]{F7FBFF}} \color[HTML]{000000} {\cellcolor{white}} 0 \\
Turkish (tur) & {\cellcolor[HTML]{F7FBFF}} \color[HTML]{000000} {\cellcolor{white}} 0 & {\cellcolor[HTML]{F7FBFF}} \color[HTML]{000000} 1 & {\cellcolor[HTML]{F7FBFF}} \color[HTML]{000000} {\cellcolor{white}} 1 & {\cellcolor[HTML]{F7FBFF}} \color[HTML]{000000} 1 & {\cellcolor[HTML]{F7FBFF}} \color[HTML]{000000} {\cellcolor{white}} 0 & {\cellcolor[HTML]{F7FBFF}} \color[HTML]{000000} {\cellcolor{white}} 0 & {\cellcolor[HTML]{F7FBFF}} \color[HTML]{000000} {\cellcolor{white}} 0 & {\cellcolor[HTML]{F7FBFF}} \color[HTML]{000000} {\cellcolor{white}} 0 & {\cellcolor[HTML]{F7FBFF}} \color[HTML]{000000} {\cellcolor{white}} 0 & {\cellcolor[HTML]{F7FBFF}} \color[HTML]{000000} {\cellcolor{white}} 0 & {\cellcolor[HTML]{DCE9F6}} \color[HTML]{000000} 140 \\
Urdu (urd) & {\cellcolor[HTML]{F7FBFF}} \color[HTML]{000000} {\cellcolor{white}} 0 & {\cellcolor[HTML]{F7FBFF}} \color[HTML]{000000} {\cellcolor{white}} 0 & {\cellcolor[HTML]{DAE8F6}} \color[HTML]{000000} 147 & {\cellcolor[HTML]{F7FBFF}} \color[HTML]{000000} {\cellcolor{white}} 0 & {\cellcolor[HTML]{F7FBFF}} \color[HTML]{000000} {\cellcolor{white}} 0 & {\cellcolor[HTML]{F7FBFF}} \color[HTML]{000000} {\cellcolor{white}} 0 & {\cellcolor[HTML]{F7FBFF}} \color[HTML]{000000} {\cellcolor{white}} 0 & {\cellcolor[HTML]{F7FBFF}} \color[HTML]{000000} {\cellcolor{white}} 0 & {\cellcolor[HTML]{F7FBFF}} \color[HTML]{000000} {\cellcolor{white}} 0 & {\cellcolor[HTML]{F7FBFF}} \color[HTML]{000000} {\cellcolor{white}} 0 & {\cellcolor[HTML]{F7FBFF}} \color[HTML]{000000} {\cellcolor{white}} 0 \\
Chinese (zho) & {\cellcolor[HTML]{F7FBFF}} \color[HTML]{000000} {\cellcolor{white}} 0 & {\cellcolor[HTML]{F7FBFF}} \color[HTML]{000000} {\cellcolor{white}} 0 & {\cellcolor[HTML]{E7F1FA}} \color[HTML]{000000} 81 & {\cellcolor[HTML]{F7FBFF}} \color[HTML]{000000} {\cellcolor{white}} 0 & {\cellcolor[HTML]{F7FBFF}} \color[HTML]{000000} {\cellcolor{white}} 0 & {\cellcolor[HTML]{F7FBFF}} \color[HTML]{000000} {\cellcolor{white}} 0 & {\cellcolor[HTML]{F7FBFF}} \color[HTML]{000000} {\cellcolor{white}} 0 & {\cellcolor[HTML]{F7FBFF}} \color[HTML]{000000} {\cellcolor{white}} 0 & {\cellcolor[HTML]{F7FBFF}} \color[HTML]{000000} {\cellcolor{white}} 0 & {\cellcolor[HTML]{F7FBFF}} \color[HTML]{000000} {\cellcolor{white}} 0 & {\cellcolor[HTML]{F7FBFF}} \color[HTML]{000000} {\cellcolor{white}} 0 \\
\bottomrule
\end{tabular}
}
\caption{Combinations of languages in SMP-Claim pairs for test set of the crosslingual track.}
\label{tab:crosslingual_test}
\end{table}

\paragraph{Test Set.}
Our test set consists of additionally collected data containing posts that are not a part of the original MultiClaim dataset and which were published until March 2024. For fact-checked claims, the test set contains the ones already present in the training and development sets, as well as newly collected ones (ranging until May 2024). The same data collection methodology was applied as in the original MultiClaim dataset, while being extended to more sources (more fact-checking organizations and also including Telegram in case of SMPs). Similarly, the same data cleaning and pre-processing was applied as described in~\citet{pikuliak-etal-2023-multilingual} to ensure consistency.

For the monolingual track, the same 8 languages were used as in the training and development sets. Additionally, we added two unannounced languages: Polish and Turkish, which both meet the criteria imposed on the number of SMP-Claim pairs and which did not appear in the prior phases of the shared task in either of the subtracks. We select a random set of 500 posts (where available; only 93 and 183 posts were available for Malay and Thai, resp.) for each language and all fact-checked claims available for that language. 

To construct a test set for the crosslingual track, we again apply the same criteria as for the training and development sets, resulting in 11 languages for claims (10 from the monolingual track plus Hindi), 14 languages in SMPs -- the ones not contained in the monolingual track being Hindi, Urdu, Chinese, and Bengali -- and 60 language combinations in total. The complete statistics for the test set are reported in Table \ref{tab:dataset_statistics}. Table~\ref{tab:crosslingual_test} shows the combinations of languages in SMP-Claim pairs contained in the test set. It is worth noting that despite the language diversity, most crosslingual pairs still contain English as either a language of a post or of a claim. This is one of the limitations of the dataset and it needs to be taken into consideration when interpreting the results of the crosslingual track.

\begin{table}[t]
\centering
\resizebox{1\columnwidth}{!}{%
\begin{tabular}{|l|p{8cm}|}
\hline
post\_id  & 27169 \\ \hline
instances & (1620128767, `fb') \\ \hline
ocr       & {[}(`Merche Gonzalez de Mingo-Sancho 1h. En mi colegio las papeletas de Vox al lado de las de Volt para liar a los abuelos... Vot vox xx', ``Merche Gonzalez de Mingo-Sancho 1 hour. In my school the Vox ballots next to Volt's to mess with the grandparents... vote vox xx'', {[}(`spa', 0.6682217717170715), (`cat', 0.2810060381889343), (`eng', 0.01799275539815426){]}){]} \\ \hline
verdicts  & Partly false information \\ \hline
text      & (`Qué casualidad, no dejan ni un cabo suelto...., cuanto más confusión crean, cuanto más caos... mayor cosecha intenta sacar la siniestra. La izquierda, siempre con la mentira y la trampa por bandera.', ``What a coincidence, they don't leave a single end loose...., the more confusion they create, the more chaos... the bigger harvest the sinister tries to get. The left, always with lies and cheating as a flag.'', {[}(`spa', 1.0){]}) \\ \hline
\end{tabular}%
}
\caption{An example of a social media post. There are five fields for each post.}
\label{tab:post_example}
\end{table}

\begin{table}
\centering
\resizebox{1\columnwidth}{!}{%
\begin{tabular}{|l|p{8cm}|}
\hline
fact\_check\_id & 74855\\ \hline
claim           & (`Jarum suntik palsu dalam sebuah video yang diklaim telah disiapkan untuk vaksinasi Covid-19 para pemimpin dunia atau elite global', `Fake syringe in a video that claims to have been prepared for the Covid-19 vaccination of world leaders or global elites', {[}(`msa', 1.0){]}) \\ \hline
instances       & {[}(1612137540, `https://cekfakta.tempo.co/fakta/1221/
keliru-jarum-suntik-palsu-di-video-ini-disiapkan-untuk-vaksinasi-covid-19-elite-global'){]} \\ \hline
title           & (`Keliru, Jarum Suntik Palsu di Video Ini Disiapkan untuk Vaksinasi Covid-19 Elite Global', `Wrong, Fake Syringe in this Video Prepared for Global Elite Covid-19 Vaccination', {[}(`msa', 1.0){]}) \\ \hline
\end{tabular}%
}
\caption{An example of a fact-checked claim. There are four fields for each claim.}
\label{tab:claim_example}
\end{table}

\paragraph{Details on Data Format.}
Each post has five fields as illustrated in Table \ref{tab:post_example}. The \textit{post\_id} refers to the id of the post in our dataset. The \textit{instances} field is used to indicate the timestamp of the post and its social platform. The \textit{ocr} field is filled with transcribed texts if there are any associated images in the post. The \textit{verdict} is assigned by professional fact-checkers. The \textit{text} field contains the content of the post, its translation to English and a list of detected languages. 

Each claim has three fields as illustrated in Table \ref{tab:claim_example}. The \textit{fact\_check\_id} is the identification of the fact-check (claim) in the dataset. The \textit{claim} field contains the content of the claim, its translation to English and a list of identified languages. Similar to the structure of posts, the \textit{instances} field is used to indicate the timestamp of the claim and its relevant URL. The \textit{title} field contains the original title of the fact-check, its translation to English and a list of identified languages. 

We utilize the Google Vision API and Google Translate API to obtain the texts in the image associated to posts and the translations of non-English texts respectively.

\subsection{Evaluation}
The participants were asked to provide top $10$ results for each SMP. The lists of fact-checked claims are provided as a search pool. We calculate \textit{success@10} to evaluate the performance of all systems: 
\[\text{success@10} = \frac{1}{N} \sum_{i=1}^{N} \mathbf{1}(\text{relevant claim in top10})\]

where \(N\) is the total number of examples (SMPs), and \(\mathbf{1}(\cdot)\) is an indicator function that returns 1 if the condition is met for the post \(i\), otherwise 0. We count a retrieval as success if at least one relevant claim is ranked in the top 10.

\section{Baselines}
We selected four approaches as baselines:

\paragraph{BM25.} BM25~\cite{robertson-zaragoza-2009-bm25} is used as a default retriever in many prior works on claim matching (see, e.g.,~\citealp{vo-lee-2020-facts, shaar-etal-2020-known, shaar-etal-2022-role}) and it also achieved competitive results especially in monolingual evaluation performed in~\citet{pikuliak-etal-2023-multilingual}. We use it with the English-translated version of the dataset.
\paragraph{GTR-T5-Large.} GTR-T5-Large~\cite{ni-etal-2022-large} was the overall best performing model in~\citet{pikuliak-etal-2023-multilingual} in monolingual as well as crosslingual settings. Since it is an English-only model, we use it with the English-translated version of the dataset. 
\paragraph{Paraphrase-Multi-v2.} We use the Paraphrase-Multilingual-MPNet-Base-v2 model, which is a part of the Sentence-BERT set of pre-trained models~\cite{reimers-gurevych-2019-sentence}. We selected this model as one of the baselines, since it was the best multilingual model in~\citet{pikuliak-etal-2023-multilingual}, although having lower performance than both BM25 and GTR-T5-Large (which, however, work on the English translations). The original multilingual version of the dataset is used with this model.
\paragraph{E5-Large.} We use Multilingual-E5-Large model~\cite{wang2024multilingual-e5}, which belongs to the best performing multilingual text embedding models under 1B parameters based on public benchmarks\footnote{\url{https://huggingface.co/spaces/mteb/leaderboard}} (2\textsuperscript{nd} for reranking task, 6\textsuperscript{th} for retrieval and sentence similarity tasks) and it had competitive results in our initial experiments. Being a multilingual model, it works with the original multilingual version of the dataset.

\begin{table*}[h!]
\centering
\resizebox{1\textwidth}{!}{%
\begin{tabular}{@{}p{7cm}|ccc|cccc|ccc@{}}
\toprule
\multirow{2}{*}{Team Name} & \multicolumn{3}{c|}{Pre-processing/Data Curation} & \multicolumn{4}{c|}{Model Training} & \multicolumn{3}{c}{Post-processing/Reranking} \\ 
& \rotatebox{45}{\parbox{2cm}{Data\\augmentation}} & \rotatebox{45}{\parbox{2.2cm}{Data filtering/ \\ cleaning}} & \rotatebox{45}{\parbox{1cm}{Other}} & \rotatebox{45}{\parbox{1.8cm}{Pre-trained\\model}} & \rotatebox{45}{\parbox{2.4cm}{Single model\\on English data}} & \rotatebox{45}{\parbox{2.5cm}{Single model\\on all languages}} & \rotatebox{45}{\parbox{2.6cm}{Ensemble of per-language models}} & \rotatebox{45}{\parbox{1.5cm}{Results\\reranking}} & \rotatebox{45}{\parbox{1.5cm}{Results\\filtering}} & \rotatebox
{45}{\parbox{1cm}{Other}} \\
\midrule

PALI                                                & & &      & \checkmark & & \checkmark &      & & & \\ \midrule
TIFIN India \citep{semeval2025task7_tifinindia}     & \checkmark & & \checkmark     & \checkmark & \checkmark & &      & \checkmark & & \\ \midrule
RACAI \citep{semeval2025task7_sherlock}          & & &      & \checkmark & & &      & & & \\ \midrule
QUST\_NLP \citep{semeval2025task7_qust_nlp}         & & & \checkmark      & \checkmark & & & \checkmark     & \checkmark & & \\ \midrule
UniBuc-AE \citep{semeval2025task7_UniBucAE}         & & &      & \checkmark & \checkmark & \checkmark &      & & & \checkmark \\ \midrule
UWBa \citep{semeval2025task7_UWBa}         & & &      & \checkmark & & &      & & & \\ \midrule
ipezoTU \citep{semeval2025task7_ipezoTU}         & & & \checkmark     & \checkmark & & &      & \checkmark & & \\ \midrule
fact check AI \citep{semeval2025task7_fc_AI}          & & \checkmark &      & \checkmark & \checkmark & \checkmark &      & & \checkmark & \\ \midrule
YNU-HPCC \citep{semeval2025task7_ynu_hpcc}         & \checkmark & \checkmark &      & \checkmark & & \checkmark &      & & & \\ \midrule
DKE-Research \citep{semeval2025task7_dke_research}         & & &      & \checkmark & & &      & & & \\ \midrule
CAIDAS \citep{semeval2025task7_caidas}         & \checkmark & &      & \checkmark & & \checkmark &      & & & \\ \midrule
UPC-HLE \citep{semeval2025task7upchle}         & & \checkmark &      & \checkmark & \checkmark & \checkmark &      & \checkmark & & \\ \midrule
ClaimCatchers \citep{semeval2025task7claimcatchers}         & & &      & \checkmark & \checkmark  & &      & \checkmark  & & \\ \midrule
Shouth NLP \citep{semeval2025task7shouth_nlp}          & & &      & \checkmark & & \checkmark &      & & & \\ \midrule
MultiMind \citep{semeval2025task7_multimind}         & \checkmark & \checkmark &      & \checkmark & & \checkmark &      & \checkmark & & \\ \midrule
JU\_NLP \citep{semeval2025task7_ju_nlp}         & & \checkmark &      & \checkmark & & &      & & & \\ \midrule
Duluth \citep{semeval2025task7duluth}         & & &      & \checkmark & & \checkmark &      & & & \\ \midrule
Howard Univeristy-AI4PC \citep{semeval2025task7howard_uni_ai4pc}         & & &      & \checkmark & & &      & & & \\ \midrule
CAISA \citep{semeval2025task7caisa}         & & \checkmark & \checkmark      & \checkmark & & \checkmark &      & & \checkmark & \\ \midrule
\end{tabular}%
}
\caption{The overview of approaches utilized in the created systems as reported by individual teams. Note: Some teams are missing as they have not responded with a report about their system. Order of teams corresponds to the rank achieved on the Monolingual Track.}
\label{tab:system_overview}
\end{table*}

\begin{table*}[h!]
\centering
\resizebox{1\textwidth}{!}{%
\begin{tabular}{@{}cp{6.8cm}ccccccccccc@{}}
\toprule
Rank & Team Name           & avg & eng & fra & deu & por & spa & tha & msa & ara & tur & pol \\ \midrule
1    & PINGAN AI           & \textbf{0.9601}     & \textbf{0.9160}     & \textbf{0.9720}     & \textbf{0.9580}     & \textbf{0.9260}     & \textbf{0.9740}     & 0.9945     & \textbf{1.0000}     & \textbf{0.9860}     & \textbf{0.9480}     & \textbf{0.9260}     \\
2    & PALI                & 0.9472     & 0.9040     & 0.9540     & 0.9360     & 0.9080     & 0.9700     & \textbf{1.0000}     & \textbf{1.0000}     & 0.9820     & 0.9300     & 0.8880     \\
3    & TIFIN India \citep{semeval2025task7_tifinindia}         & 0.9383     & 0.8800     & 0.9540     & 0.9360     & 0.9020     & 0.9600     & 0.9945     & \textbf{1.0000}     & 0.9660     & 0.9040     & 0.8860     \\
4    & RACAI \citep{semeval2025task7_sherlock}            & 0.9377     & 0.9000     & 0.9420     & 0.9280     & 0.8960     & 0.9520     & 0.9945     & \textbf{1.0000}     & 0.9660     & 0.9160     & 0.8820     \\
5    & QUST\_NLP \citep{semeval2025task7_qust_nlp}            & 0.9365     & 0.8940     & 0.9500     & 0.9020     & 0.8900     & 0.9480     & 0.9945     & \textbf{1.0000}     & 0.9700     & 0.9300     & 0.8860     \\
6    & UniBuc-AE \citep{semeval2025task7_UniBucAE}           & 0.9336     & 0.8860     & 0.9200     & 0.9320     & 0.8800     & 0.9620     & \textbf{1.0000}     & \textbf{1.0000}     & 0.9700     & 0.9100     & 0.8760     \\
7    & UWBa \citep{semeval2025task7_UWBa}               & 0.9270     & 0.8800     & 0.9440     & 0.9160     & 0.8540     & 0.9380     & \textbf{1.0000}     & \textbf{1.0000}     & 0.9540     & 0.9120     & 0.8720     \\
8    & ipezoTU \citep{semeval2025task7_ipezoTU}            & 0.9259     & 0.8900     & 0.9440     & 0.9180     & 0.8800     & 0.9300     & 0.9836     & 0.9892     & 0.9400     & 0.9100     & 0.8740     \\
9    & kubapok             & 0.9245     & 0.8700     & 0.9420     & 0.9220     & 0.8680     & 0.9420     & 0.9945     & 0.9785     & 0.9440     & 0.9120     & 0.8720     \\
10   & fact check AI \citep{semeval2025task7_fc_AI}    & 0.9232     & 0.8820     & 0.9440     & 0.9260     & 0.8660     & 0.9420     & 0.9945     & 0.9892     & 0.9400     & 0.8840     & 0.8640     \\
11   & YNU-HPCC \citep{semeval2025task7_ynu_hpcc}            & 0.9218     & 0.8520     & 0.9540     & 0.9040     & 0.8740     & 0.9400     & 0.9727     & 0.9892     & 0.9580     & 0.8960     & 0.8780     \\
12   & UWOB                & 0.9190     & 0.8800     & 0.9340     & 0.9060     & 0.8540     & 0.9380     & 0.9781     & 0.9677     & 0.9540     & 0.9060     & 0.8720     \\
13   & TM\_Trek            & 0.9189     & 0.8440     & 0.9380     & 0.9180     & 0.8180     & 0.9480     & 0.9945     & \textbf{1.0000}     & 0.9660     & 0.8840     & 0.8780     \\
14   & joeblack            & 0.9105     & 0.8160     & 0.9280     & 0.9160     & 0.8040     & 0.9340     & 0.9891     & \textbf{1.0000}     & 0.9620     & 0.8800     & 0.8760     \\
15   & DKE-Research \citep{semeval2025task7_dke_research}       & 0.8979     & 0.8200     & 0.9240     & 0.8680     & 0.8340     & 0.9160     & 0.9508     & \textbf{1.0000}     & 0.9360     & 0.8740     & 0.8560     \\
16   & CAIDAS \citep{semeval2025task7_caidas}              & 0.8953     & 0.8340     & 0.9100     & 0.9000     & 0.8340     & 0.8860     & 0.9836     & 0.9677     & 0.9340     & 0.8680     & 0.8360     \\
17   & UPC-HLE \citep{semeval2025task7upchle}             & 0.8915     & 0.7940     & 0.9460     & 0.9220     & 0.8340     & 0.9140     & 0.9781     & 0.9892     & 0.9460     & 0.7920     & 0.8000     \\
18   & ClaimCatchers \citep{semeval2025task7claimcatchers}       & 0.8780     & 0.8100     & 0.9100     & 0.8420     & 0.8000     & 0.8860     & 0.9727     & 0.9570     & 0.9320     & 0.8740     & 0.7960     \\
19   & NCL-AR \citep{semeval2025task7ncl_ar}              & 0.8716     & 0.8340     & 0.9180     & 0.8980     & 0.7980     & 0.8780     & 0.9454     & 0.9462     & 0.8840     & 0.8140     & 0.8000     \\
20   & Shouth NLP \citep{semeval2025task7shouth_nlp}           & 0.8674     & 0.7960     & 0.8940     & 0.8580     & 0.7960     & 0.8660     & 0.9399     & 0.9785     & 0.9120     & 0.8280     & 0.8060     \\
\rowcolor[HTML]{C0C0C0} 
*    & E5-Large            & 0.8589     & 0.8180     & 0.8540     & 0.8720     & 0.8080     & 0.8560     & 0.9290     & 0.8602     & 0.9160     & 0.8780     & 0.7980     \\
\rowcolor[HTML]{C0C0C0} 
*    & BM25                & 0.8123     & 0.7500     & 0.8220     & 0.8120     & 0.7540     & 0.7960     & 0.9071     & 0.9140     & 0.8460     & 0.7900     & 0.7320     \\
21   & MultiMind \citep{semeval2025task7_multimind}           & 0.8080     & 0.6740     & 0.8640     & 0.8000     & 0.7480     & 0.7760     & 0.9235     & 0.9570     & 0.8480     & 0.7460     & 0.7440     \\
    22   & JU\_NLP \citep{semeval2025task7_ju_nlp}             & 0.7868     & 0.6540     & 0.8700     & 0.7320     & 0.6460     & 0.6840     & 0.9290     & 0.9570     & 0.8740     & 0.7800     & 0.7420     \\
\rowcolor[HTML]{C0C0C0} 
*    & GTR-T5-Large        & 0.7299     & 0.7880     & 0.7300     & 0.7380     & 0.7140     & 0.7320     & 0.7049     & 0.7097     & 0.8620     & 0.7160     & 0.6040     \\
23   & Duluth \citep{semeval2025task7duluth}              & 0.6883     & 0.4520     & 0.8140     & 0.6900     & 0.5580     & 0.5460     & 0.8415     & 0.8495     & 0.8200     & 0.6860     & 0.6260     \\
\rowcolor[HTML]{C0C0C0} 
*    & Paraphrase-Multi-v2 & 0.5683     & 0.6180     & 0.6040     & 0.5340     & 0.4840     & 0.5160     & 0.6175     & 0.5054     & 0.6940     & 0.5740     & 0.5360     \\
24   & Word2winners \citep{semeval2025task7Word2winners}        & 0.5488     & 0.6160     & 0.5460     & 0.5320     & 0.6000     & 0.5480     & 0.4372     & 0.4731     & 0.7080     & 0.5220     & 0.5060     \\
25   & UMUTeam \citep{semeval2025task7umuteam}             & 0.5445     & 0.4140     & 0.5640     & 0.3840     & 0.4700     & 0.4200     & 0.7869     & 0.6882     & 0.7160     & 0.5380     & 0.4640     \\
26   & FactDebug \citep{semeval2025task7fact_debug}           & 0.2213     & 0.3620     & 0.2640     & 0.4240     & 0.2360     & 0.1820     & 0.2350     & 0.0645     & 0.4460     & 0.0000     & 0.0000     \\
27   & TransformerHHU      & 0.1499     & 0.1980     & 0.1080     & 0.1800     & 0.1880     & 0.2220     & 0.1148     & 0.2043     & 0.0660     & 0.1140     & 0.1040     \\ \bottomrule
\end{tabular}%
}
\caption{The results (success@10) on Monolingual Track. Teams are ranked by their average performance across all languages. The best results for each language and the average are in bold. The rows with gray background and * indicate the performance of four baselines. Teams with reference have submitted system papers. }
\label{tab:monolingual_track_results}
\end{table*}

\section{Participating Systems and Results}

The high-level overview of approaches utilized by individual systems is summarized in Table \ref{tab:system_overview}. The results presented in Tables \ref{tab:monolingual_track_results} and \ref{tab:crosslingual_track_results} reflect the performance of test submissions by the end of the test phase. In their system papers, participants may provide additional post-test analysis that achieve further improvements.

\subsection{Monolingual Track}
In the monolingual track, posts and claims are in the same language, and multilingual data cover ten different languages. Two (Turkish and Polish) out of ten are unannounced languages -- i.e., no examples in these languages appear in the training and development sets. 

21 out of 27 teams have submitted their system description papers on the monolingual track. The results are summarized in Table \ref{tab:monolingual_track_results}. Out of all teams that submitted system description papers, we describe the top-5 performing systems.  

\paragraph{TIFIN India.} Inspired by \citet{khan2024indicllmsuite}, \citet{semeval2025task7_tifinindia} first translate all posts and claims into English and utilize English-focused models for the claim retrieval task. They employ the Aya-expanse-8B \citep{dang2024aya} model for the translation. Then, a two-stage pipeline is implemented. The Stella 400M embedding model \citep{zhang2024jasper} is first used to retrieve top 50 claims using cosine similarity. These are fed into an LLM-based re-ranker, Qwen2.5-72B-Instruct \citep{qwen2.5}. The re-ranking is performed in a prompting style. The top 10 candidates from the re-ranker are then combined with the top 10 results from the embedding model by Reciprocal Rank Fusion \citep{cormack2009reciprocal}, producing the final top 10 results. They further fine-tune the embedding model by adding hard negatives, demonstrating that 20 negatives per post give the best performance. 

\paragraph{RACAI.} \citet{semeval2025task7_sherlock} adopt a single-step strategy to tackle the task without re-ranking. All posts and claims are used in their original languages. The BGE-Multilingual-Gemma2-9B model \citep{bge_embedding, bge-m3}, is used to provide embeddings for posts and claims. They explored different strategies to optimize and adapt the general-purpose embedding model on the given task. They first utilize LoRA \cite{hu2022lora} with a contrastive learning objective with in-batch negatives to fine-tune the base model. Additionally, they use prompt-tuning to adjust the embeddings for instructions. To reduce the computational requirements, Matryoshka representation learning \citep{kusupati2022matryoshka} is employed to produce embeddings of different dimension sizes.

\paragraph{QUST\_NLP.} \citet{semeval2025task7_qust_nlp} propose a three-stage retrieval framework. A group of six different retrieval models 
is utilized in the retrieval stage to coarsely identify relevant claims for each post. In the re-ranking stage, six different re-ranking models 
are employed to provide top 10 candidates from the retrieved results. For each language, a set of re-ranker models are fine-tuned with training data in that language. Finally, the weighted voting stage combines the top 10 candidates from each re-ranker and weight them by their performance on the dev set to obtain the final top 10 results. They also find that the combination of the original text and corresponding English translations can further improve the overall performance.    

\paragraph{UniBuc-AE.} \citet{semeval2025task7_UniBucAE} utilize both the original text and English-translated text for retrieval. Two embedding models (English one: e5-large-v2, \citealp{wang2022text}, and multilingual one: multilingual-e5-large-instruct, \citealp{wang2024multilingual}) are employed to provide corresponding embeddings after contrastive learning with in-batch negatives. They further fine-tune the two models with hard negatives, resulting in two hard fine-tuned models. All four models are used to vote for the final top 10 retrieval claims in a weighted manner. The optimal weights are discovered by their performance on the development set. 

\paragraph{UWBa.} \citet{semeval2025task7_UWBa} present a zero-shot claim retrieval approach with all posts and claims translated into English. Five embedding models without any further fine-tuning are utilized to provide embeddings for posts/claims. A model combination strategy is employed to produce the final top 10 candidates. The models are firstly ranked in term of performance on dev set. Their top 5 retrieval results are then combined to produce the top 10 results after potential de-duplication. 

\paragraph{Discussion.} We can see that most systems outperformed the baselines. On the other hand, a clear pattern is that the two unseen languages, Turkish and Polish, show a generally worse performance than the languages observed in the training data. This indicates that the \textbf{generalization ability to unseen languages is still a big challenge}. 

Approximately half of the submissions to the monolingual track did not apply any data filtering or pre-processing techniques, including many good-performing competitors. This does not mean that these techniques (e.g., using LLMs for data augmentation) do not improve performance, but others, such as fine-tuning, have a higher impact.  On the other hand, a common and successful strategy (used by three out of top-5 described systems) was to improve base embedding models by \textbf{fine-tuning them in a contrastive learning manner} with in-batch negatives and using hard-negatives for further improvements. 

Many systems utilized a one-stage approach (retrieval with fine-tuned or vanilla embedding models) to tackle the retrieval task, while some systems built two- or three-stage pipelines (adding a re-ranking or a weighted voting stage) that demonstrated a better performance in some cases. However, the results do not provide a conclusive evidence whether adding them is in general better, especially since we do not compare the systems in terms of their computational requirements. 

In the monolingual track, there are subsets for different languages. Yet, most systems chose to train one model for all languages. This was achieved by either translating everything into English and utilizing an English-focused model, or, more often, utilizing a multilingual embedding model with the original data. When comparing the former (e.g., TIFIN India) with the latter (e.g., RACAI), we can see that \textbf{the differences between the two approaches are negligible, which clearly shows the recent improvements in multilingual models} compared to the situation reported in~\citet{pikuliak-etal-2023-multilingual}. A small number of systems trained a set of models for each language specifically. Some other models utilized both texts in original language and their English translations, reporting improved performance. 


\begin{table}[t]
\centering
\resizebox{1\columnwidth}{!}{%
\begin{tabular}{@{}clc@{}}
\toprule
Rank & Team Name                 & S@10 \\ \midrule
1    & PINGAN AI                 & \textbf{0.85875}    \\
2    & PALI                      & 0.82675    \\
3    & RACAI \citep{semeval2025task7_sherlock}                 & 0.82450    \\
4    & TIFIN India \citep{semeval2025task7_tifinindia}              & 0.81025    \\
5    & fact check AI \citep{semeval2025task7_fc_AI}          & 0.79750    \\
6    & kubapok                   & 0.79700    \\
7    & QUST\_NLP \citep{semeval2025task7_qust_nlp}                  & 0.79250    \\
8    & UniBuc-AE \citep{semeval2025task7_UniBucAE}                 & 0.79000    \\
9    & UWBa \citep{semeval2025task7_UWBa}                      & 0.78250    \\
10   & UWOB                      & 0.78250    \\
11   & YNU-HPCC \citep{semeval2025task7_ynu_hpcc}                 & 0.77025    \\
12   & ipezoTU \citep{semeval2025task7_ipezoTU}                  & 0.74775    \\
13   & CAIDAS \citep{semeval2025task7_caidas}                   & 0.74475    \\
14   & TM\_Trek                  & 0.73975    \\
15   & ClaimCatchers \citep{semeval2025task7claimcatchers}             & 0.73675    \\
16   & joeblack                  & 0.71875    \\
17   & DKE-Research \citep{semeval2025task7_dke_research}              & 0.71325    \\
\rowcolor[HTML]{C0C0C0} 
*    & GTR-T5-Large              & 0.70525    \\
18   & Team I2R                  & 0.69725    \\
19   & UPC-HLE \citep{semeval2025task7upchle}                   & 0.63850    \\
\rowcolor[HTML]{C0C0C0} 
*    & E5-Large                  & 0.63725    \\
20   & JU\_NLP \citep{semeval2025task7_ju_nlp}                   & 0.61900    \\
\rowcolor[HTML]{C0C0C0} 
*    & BM25                      & 0.60275    \\
21   & Howard Univeristy-AI4PC \citep{semeval2025task7howard_uni_ai4pc} & 0.59225    \\
22   & Word2winners \citep{semeval2025task7Word2winners}             & 0.55425    \\
23   & CAISA \citep{semeval2025task7caisa} & 0.54475    \\
24   & MultiMind \citep{semeval2025task7_multimind}                & 0.48900    \\
\rowcolor[HTML]{C0C0C0} 
*    & Paraphrase-Multi-v2       & 0.41075    \\
25   & NCL-AR \citep{semeval2025task7ncl_ar}                    & 0.39775    \\
26   & FactDebug \citep{semeval2025task7fact_debug}                 & 0.32000    \\
27   & UMUTeam \citep{semeval2025task7umuteam}                   & 0.28025    \\
28   & DANGNT-SGU                & 0.00025    \\ \bottomrule
\end{tabular}%
}
\caption{The results (success@10) on Crosslingual Track. Teams are ranked by their performance. The best result is in bold. The rows with gray background and * indicate the performance of baselines. Teams with reference have submitted system papers.}
\label{tab:crosslingual_track_results}
\end{table}

\subsection{Crosslingual Track}
20 out of 28 teams submitted system description papers for the crosslingual track. The results are summarized in Table \ref{tab:crosslingual_track_results}. We describe the top 5 performing systems of those teams that submitted system description papers.

\paragraph{RACAI.} The same pipeline is employed as for the monolingual track, where the base embedding model is improved by contrastive learning with in-batch negatives and prompt-tuning. They demonstrate that it brings much higher improvements in crosslingual compared to monolingual setup.  

\paragraph{TIFIN India.} They translate everything into English and use the same two-stage pipeline as in the monolingual track. Their high performance shows that this strategy is effective in both monolingual and crosslingual setups. 

\paragraph{fact check AI.} \citet{semeval2025task7_fc_AI} uses the English translations. Base embedding models (mul-e5-large-instruct, Stella-400M, and mxbai-embed-large-v1) are fine-tuned with contrastive learning. The fine-tuning is performed in a K-fold manner to increase the robustness and prevent over-fitting. The final 10 candidates are produced by models' voting. They also conduct pre-processing such as removing noise (e.g., url, emoji, extra space) and filtering out too-short texts.  

\paragraph{QUST\_NLP.} The same three-stage framework is utilized for the crosslingual track. Different from the language-specific fine-tuning strategy for re-rankers in the monolingual track, they use English translations and employ one set of re-rankers for refined re-ranking. They also demonstrate that though the combination of the original text and English translations are helpful in the monolingual track, such combination often causes performance decrease in the crosslingual track.        

\paragraph{UniBuc-AE.} They adopt the same strategy for the crosslingual track where four embedding models are used for weighted voting. They show that fine-tuning with hard negatives brings larger improvements in the crosslingual setup.

\paragraph{Discussion.} We can see that most systems outperform the baselines, but \textbf{they achieve a worse performance compared to the monolingual track (more than 10 percentage points difference)}, highlighting the challenge of this setup. Also, most systems that were submitted to both tracks do not consider a new approach but adopt the same pipeline, which seems to work quite well, given that it is the case for all top-5 best performing systems in the crosslingual track. However, they also highlight some performance inconsistencies of different techniques when applied across the tracks. For example, the \textbf{improvements brought in by additional re-ranking and weighted-voting seem to be larger} in the crosslingual setup. On the other hand, as indicated by \citet{semeval2025task7_qust_nlp}, the combination of original text with English translations instead has a negative impact on the performance in the crosslingual setup. This raises questions on the transferability of such techniques and models to the crosslingual scenario, which a future research should examine more comprehensively.

\section{Conclusion}
This work presents an overview of SemEval 2025 Task 7 on multilingual and crosslingual fact-checked claim retrieval. It provides a comprehensive analysis of the dataset and submitted systems. The task has attracted 179 participants. In the final test phase, 31 teams submitted test systems, of which 23 submitted system description papers. 

Our analysis reveals that the most common strategy to improve base embedding models is to fine-tune them in a contrastive learning manner with in-batch negatives. As demonstrated by many systems, further improvements can be obtained by adding hard negatives and training on diverse subsets. A multi-stage pipeline is adopted by a large number of systems that additionally include a re-ranking and a weighted voting stage, which can bring improvements in both crosslingual and monolingual setups. An effective strategy to tackle different languages is still to translate them into English and utilize an English-focused model, but using multilingual models with the texts in original languages already gives comparable results. Several systems have discovered that some techniques which work well in one track have a different impact in the other track, suggesting the difficulty of transferability. 

We believe our shared task along with participating systems provides valuable insights into multilingual and crosslingual claim retrieval, supporting future research in this field.


\section*{Ethical Considerations}

Most of the ethical, legal and societal issues tied to the MultiClaim dataset were already described in the Ethical Considerations section of the accompanying paper \cite{pikuliak-etal-2023-multilingual}. The most severe risks were tied to a Terms of Service (ToS) violation, various types of privacy intrusions, the possibility of third-party misuse, or the erosion of some privacy rights such as the right to erasure. For the shared task organization we also discussed the possibility of the emergence of new issues that were not assessed before with respect to the most affected stakeholders, especially researchers who will participate in the shared task, social media users, and social media platforms.

We have reassessed the risk of a potential violation of the ToS of the social media platforms in light of the new EU digital regulations. Exploratory research on very large online platforms is now legally permitted by Article 40 (12) of the EU Act on digital services (DSA) if the research concerns systemic risks. As the spread of disinformation is clearly a systemic risk as foreseen by Recital 83 of the DSA \cite{DSA_law}, we see this as an argument in favor of the further use of the MultiClaim dataset.   

During the shared task, new methods were proposed, trained and published by the participants. As strictly research-oriented models, they should not be deployed without further assessment regarding their potential misuse, privacy concerns, or the occurrence of various forms of algorithmic biases \cite{lee_risk_identification}. To avoid the risk that participants do not understand the limitations and further uses of data and methods used in our shared task, we prepared a set of guidelines to inform participants from the beginning of the task about the purpose of the task, its possible limitations, and the admissible uses of the outputs,

In the process of analyzing data, researchers participating in our shared task may have been exposed to several risks tied to their well-being, moral integrity, or safety. 
Disinformation may contain some sensitive societal topics regarding the LGBTQIA+ community, war, or humanitarian crises, child abuse, terrorism, or other political and theological topics. However, since the participants focused on improving retrieval performance rather than direct content analysis of the posts, we expect possible negative impacts to be minimal. 

To minimize the risks of third-party misuse or revealing incorrect, highly sensitive, or offensive content, we still maintain the right to restrict the use of the shared task dataset. 
Participants were not allowed to use any external datasets other than the dataset prepared for the shared task, to enable fair evaluation. To ensure that any residual privacy concerns are adequately addressed, contact persons were designated, through which participants were able flag potential data issues. 

\section*{Acknowledgement}
This work was supported by DisAI - Improving scientific excellence and creativity in combating disinformation with artificial intelligence and language technologies, a project funded by European Union under the Horizon Europe, GA No. \href{https://doi.org/10.3030/101079164}{101079164}.

\balance
\bibliography{anthology,custom}
\bibliographystyle{acl_natbib}

\appendix

\end{document}